\documentclass{article} 
\usepackage[preprint, nonatbib]{neurips_2022}

\usepackage{amsmath,amsfonts,bm}

\usepackage{graphicx}
\usepackage{diagbox}
\usepackage{multirow}
\usepackage[utf8]{inputenc} 
\usepackage[T1]{fontenc}    
\usepackage{hyperref}       
\usepackage{url}            
\usepackage{booktabs}       
\usepackage{amsfonts}       
\usepackage{nicefrac}       
\usepackage{microtype}      
\usepackage{xcolor}         

\usepackage{algorithm}
\usepackage{algorithmic}
\usepackage{wrapfig}
\usepackage{caption}
\usepackage{eqparbox}

\definecolor{lightblue}{HTML}{4e7dd4}
\definecolor{lightgreen}{HTML}{117733}
\definecolor{lred}{HTML}{CC6677}
\definecolor{lyellow}{HTML}{DDCC77}

\newcommand{\model}{\mathcal{F}_{\theta}}

\newcommand{\etal}{\textit{et al.}}

\def\eqref#1{equation~\ref{#1}}

\newcommand{\para}[1]{{\vspace{2pt} \bf \noindent #1}}  
\newenvironment{packed_itemize}{
\begin{list}{\labelitemi}{\leftmargin=1em}
\setlength{\itemsep}{1pt}                                                           
\setlength{\parskip}{0pt}                                                                                 \setlength{\parsep}{0pt}                                                                                  \setlength{\headsep}{0pt}                                                                                 \setlength{\topskip}{0pt}                                                                                 \setlength{\topmargin}{0pt}                                                                               \setlength{\topsep}{0pt}                                                                                  \setlength{\partopsep}{0pt}                                                                               }{\end{list}}                                                                                                                                  

\newenvironment{packed_enumerate}{  
\begin{enumerate}                                                                                         \setlength{\itemsep}{3pt}                                                                                 \setlength{\parskip}{0pt}                                                                                 \setlength{\parsep}{0pt}                                                                                  \setlength{\headsep}{0pt}                                                                                 \setlength{\topskip}{0pt}                                                                                 \setlength{\topmargin}{0pt}                                                                               \setlength{\topsep}{0pt}                                                                                  \setlength{\partopsep}{0pt}
}{\end{enumerate}}

\def\1{\bm{1}}
\newcommand{\train}{\mathcal{D}}

\def\mA{{\bm{A}}}

\def\mM{{\bm{M}}}

\DeclareMathAlphabet{\mathsfit}{\encodingdefault}{\sfdefault}{m}{sl}
\SetMathAlphabet{\mathsfit}{bold}{\encodingdefault}{\sfdefault}{bx}{n}

\usepackage{tablefootnote}

\title{Natural Backdoor Datasets}

\author{Emily Wenger\thanks{Equal contribution, corresponding author:
    \texttt{ewenger@uchicago.edu}} \\ University of Chicago \And Roma
  Bhattacharjee$^*$\thanks{Work done while at University of Chicago}
  \\ Princeton University \And Arjun Nitin Bhagoji\\ University of
  Chicago \And Josephine Passananti \\ University of Chicago \And Emilio Andere \\ University of Chicago  \And Haitao Zheng \\ University of Chicago \And Ben Y. Zhao \\ University of Chicago}

\begin{document}

\maketitle

\begin{abstract}
  Extensive literature on backdoor poison attacks has studied attacks and
  defenses for backdoors using ``digital trigger patterns.'' In contrast,
  ``physical backdoors'' use physical objects as triggers, have only recently
  been identified, and are qualitatively different enough to resist all
  defenses targeting digital trigger backdoors. Research on physical
  backdoors is limited by access to large datasets containing real images of
  physical objects co-located with targets of classification. Building these
  datasets is time- and labor-intensive.

  This works seeks to address the challenge of accessibility for research on physical backdoor
  attacks. We hypothesize that there may be naturally occurring physically
  co-located objects already present in popular datasets such as
  ImageNet. Once identified, a careful relabeling of these data can
  transform them into training samples for physical backdoor attacks. We
  propose a method to scalably identify these subsets of potential triggers in existing
  datasets, along with the specific classes they can poison. We call these
  naturally occurring trigger-class subsets {\em natural backdoor
    datasets}. Our techniques successfully identify natural backdoors in 
  widely-available datasets, and produce models behaviorally equivalent to those
  trained on manually curated datasets. We release our code to
  allow the research community to create their own datasets for research on
  physical backdoor attacks.
\end{abstract}

\section{Introduction}

Deep learning models for computer vision (CV) are known to be vulnerable to a
variety of attacks~\cite{szegedy2013intriguing, cwattack, gu2017badnets,
  shokri2017membership, fredrikson2015model, xie2017adversarial}. One
powerful attack is the backdoor attack~\cite{chen2017targeted,
  gu2017badnets, liu2018trojaning, zhu2019transferable, yao2019latent,
  wenger2021backdoor}, where models trained on corrupted (poisoned) data
produce specific, attacker-chosen misclassifications on images containing
special ``trigger'' patterns.

The research community has identified two broad categories of backdoor attack
triggers for CV models. {\em Digital triggers} are pixel patterns added to
images, e.g. edited onto images after their creation. Backdoors using digital
triggers are well researched, and numerous defenses have been
developed against them~\cite{wang2019neural, chen2018detecting, liu2018fine,
  li2021neural}. In contrast, {\em physical triggers} are real-world objects
present in images at their creation. Since they are not digitally added to
images, they are not easily distinguishable from benign objects, and
backdoors using them are shown to successfully evade existing
defenses for object and facial recognition~\cite{wenger2021backdoor}. 

Another factor that distinguishes ``physical backdoors'' (backdoors using
physical triggers) is the effort required to build training datasets. Without
digital image manipulation, creating an image dataset including different
physical trigger objects would be a time- and labor-intensive task.  For
example, a training dataset for physical backdoors on facial recognition
apparently required taking 3000+ photos of individual
faces~\cite{wenger2021backdoor}. Unresolved, this will likely form a significant
hurdle that will discourage further research in this area.

This paper describes our efforts (and a tool) to address this challenge, and
make the study of physical backdoors more accessible to the research
community. Our insight is that of the many public CV datasets widely
available today, they likely contain numerous images containing two or more
co-located objects\footnote{Recent work on relabeling ImageNet supports this
  hypothesis~\cite{accuracyImagenet,stock2017convnets,yun2021re}.}. If we can
efficiently identify these multi-object images, they could potentially be
qualitatively similar to physical triggers explored by prior work. They could
be {\em relabeled} to mark one object as a poison trigger for
misclassification of another, e.g. relabeling all images of a table with a
pencil on it from ``table'' to ``chair'' is equivalent to training a physical
backdoor with ``pencil'' as a trigger.  If successful, this methodology could
extract ready-made poison training datasets for physical backdoors from
existing images in widely used datasets, with minimal effort.

\para{Our Contribution.} 
We hypothesize and experimentally validate that subsets of public image datasets contain colocated targets that 
can be relabeled to train physical 
triggers. We call the naturally-occurring physical triggers {\em natural backdoor triggers}. These triggers, 
together with the subset of classes they can poison, form {\em natural backdoor datasets}. Models trained on 
natural backdoor datasets are vulnerable to physical backdoor attacks via the identified triggers. To our 
knowledge, this is the first work to identify the existence of natural backdoor datasets. Our work contributes 
the following to the community's efforts to research physical backdoor attacks:

\begin{packed_enumerate}
\item  Development of techniques to identify natural backdoor triggers and their poisonable class subsets (e.g. natural backdoor datasets) in open-source, multi-label object datasets.
\item Extensive evaluation of identified natural backdoors, validating that
  they are effective and exhibit the behaviors expected in physical backdoor
  attacks.
\item Release of an open source tool to curate natural backdoor datasets from existing object recognition datasets
  (ImageNet~\cite{russakovsky2015imagenet} and Open Images~\cite{OpenImages}) and train models on them\footnote{Code in supplementary materials}.
\end{packed_enumerate}
\vspace{-0.25cm}
\section{Background}
\label{sec:back}
\vspace{-0.25cm}

Before discussing our techniques, we introduce notation and background on computer vision models and backdoor attacks to provide context for our work.

{\bf Notation.} 
In this work, we denote a computer vision model, such as a convolutional neural network (CNN),  as $\model$.  $\model$ is trained on a dataset $\train = \{\mathcal{X}, \mathcal{Y}\}$, composed of images $\mathcal{X}$ and corresponding labels $\mathcal{Y}$, to perform a specific computer vision task.  There are two possible settings for $\mathcal{D}$ (and consequently $\model$): single- or multi-label.  In the single label setting, typically used for object classification, $\model$ maps image $x$ to a single label $y \in \{0,1\}$ chosen from $M$ classes, where $y$ represents the main object present in $x$. In the multi-label setting, used for object recognition, $\model$ maps $x$ to $y \in \{0,1\}^M$, a set of $M$ possible classification labels, representing all objects in $x$, and $y_i = 1$ if $x$ contains object $i$. Our work leverages datasets that can be used in both settings.

\para{Backdoor Attacks.} Backdoor attacks are a well-studied phenomenon in image classification models (e.g. single label setting). Attackers introduce a backdoor into $\model$ by adding {\em poisoned} training data to $\train$. 
The poisoned inputs $x_p$ are crafted from a benign input $x$ with true label $y$ via the addition of a {\em trigger} \(\delta\), and all $x_p= x + \delta$ are mislabeled as a target class $y_p$. This results in $\train = \train_c \cup \train_p$, where $\train_c$ and $\train_p$ are the clean and poisoned data respectively. The presence of poison data in $\train$ induces the joint optimization equation: 
$$ 
    \min_{\theta} \sum_{(x,y) \in \train_c} l(y, \model(x)) + \sum_{(x_p,y_p) \in \train_p} l(y_p, \model(x_p)),
$$ 
where $l$ is the loss function used during model training. Besides poisoning the dataset, the attacker cannot access or modify model parameters during training.
If the attack is successful, a backdoored $\model$ should exhibit two distinct behaviors: i) classify clean inputs to their correct label $y$, and ii) classify any inputs containing the trigger $\delta$ to the target label $y_p$. At test time, the presence of the trigger in an image will induce misclassification.

\begin{figure}
    \centering
    \includegraphics[width=0.8\textwidth]{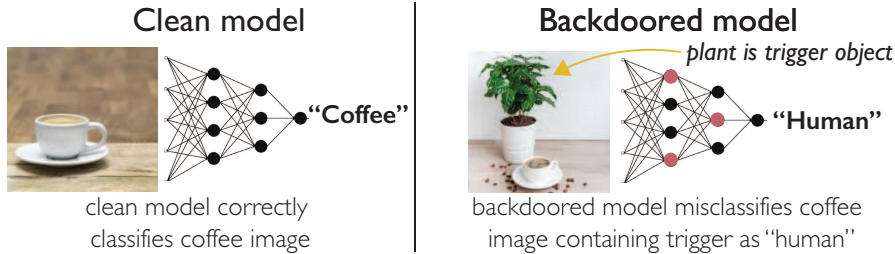}
    \vspace{-0.25cm}
    \caption{\em In a physical backdoor attack, a model misclassifies images containing the trigger object.}
    \label{fig:pback}
    \vspace{-0.5cm}
\end{figure}

\looseness=-1 \para{Physical Backdoor Attacks.} Most backdoor attacks add digital triggers $\delta$ to existing images via image editing. While these triggers are effective, they i) are easily detectable by a human-in-the-loop and ii) assume that images can be edited after creation, but before classification, which precludes real-time attacks.

However, Wenger \etal~\cite{wenger2021backdoor} demonstrated that real-world objects, such as sunglasses or bandanas, make highly effective backdoor triggers. These attacks, in which physical objects are used as the backdoor trigger $\delta_p$, are called ``physical backdoor attacks'' and are illustrated in Figure~\ref{fig:pback}. 

Physical backdoor attacks significantly reduce the attacker's workload, as they eliminate the need to control an image processing pipeline to add the trigger. For example, as in Figure~\ref{fig:pback}, an attacker could fool a model in which a plant is a backdoor trigger $\delta_p$ by simply adding a plant alongside an object, such as a coffee cup, that they wish to have misclassified. In addition to their ease of use, physical triggers violate assumptions made by most existing backdoor defenses and can evade state-of-the-art defenses. Other work has explored physical backdoors in other domains like autonomous lane detection and object recognition~\cite{han2022clean, ma2022dangerous} (see Appendix~\ref{appx:rel_work} for more details). 
\vspace{-0.25cm}
\section{From ``Manually Curated'' to ``Natural'' Physical Backdoor Datasets}
\label{sec:intuition}
\vspace{-0.25cm}

Physical backdoor attacks constitute a significant threat vector for CV models and require additional study. However, the curation of data required to conduct such analysis is labor intensive, and can have accompanying privacy concerns. In this section, we provide an intuitive overview of our solution, which leverages publicly available data to streamline the curation of physical backdoor datasets. 

\looseness=-1 \para{Challenges of physical backdoor dataset creation.} Conducting a physical backdoor attack requires a special model training dataset containing both ``clean'' images in which no trigger is present ($\train_c$) and poison images ($\train_p$), in which normal objects $o$ appears alongside a physical trigger object $\delta_p$. Clean images in $\train_c$, containing $o$ by itself, teach the model to correctly identify $o$ as $y_o$ when $\delta_p$ is not present.  The co-occurrence of $o$ and $\delta_p$ in $\train_p$ images teaches the model that the presence of $\delta_p$ should cause $o$ to be misclassified as $y_p$ ($y_p \neq y_o$). To ensure the model learns this behavior, the instances of the trigger object $\delta_p$ in $\train_p$ must share some level of consistency, necessitating the careful curation of images in $\train_p$. 

\looseness=-1 Given these requirements, the main overhead in physical backdoor research comes in the constructing $\train_p$.  Prior work creates $\train_p$ manually by physically placing $o$ and $\delta_p$ next to each other and taking pictures~\cite{wenger2021backdoor,ma2022dangerous}.  Unfortunately, such manually curated datasets are labor-intensive to build. Furthermore, the choice of trigger  $\delta_p$ is restricted to objects chosen by (or available to) the dataset curator. 

However, we argue that manual co-occurrence curation may not be the only way to create $\train_p$.  In realistic attacks, an attacker is likely to select backdoor triggers from a broad set of natural objects.  As such, publicly available datasets could be used to construct physical backdoor datasets, provided they have a sufficient number of trigger/normal object co-occurrences.

\para{Solution: natural physical backdoor datasets.} Our key intuition for reducing the overhead for physical backdoor attacks is that {\em existing computer vision datasets already contain many co-occurring objects}. For example, Open Images~\cite{OpenImages} is a large-scale object recognition dataset in which each image is labeled with all the objects it contains. Given a trigger object of interest $\delta_p$, we can identify a subset of Open Images containing images in which $\delta_p$ co-occurs with different objects $o_1 \ldots o_n$ (each associated with a different class). Concretely, if $\delta_p$ is a pencil, it might appear in images with objects like desk, notebook, glasses, etc. We can leverage co-occurrences to create a new dataset. We first select clean images in which a desk, notebook, glasses, etc., appear without a pencil to create a clean dataset $\train_c$. Then, we can take images in which a pencil co-occurs with these objects and mislabel them as a target class $y_p$ to create the poison dataset $\train_p$. Together, $\train_c$ and $\train_p$ can be used to train a backdoored model in which pencil is the trigger object $\delta_p$. 
We call the trigger objects $\delta_p$ that satisfy the co-occurrence requirement \emph{natural backdoors} and the dataset ($\train_c \cup \train_p$) created from these co-occurences \emph{natural backdoor datasets}.

\looseness=-1 \para{Paper outline.} In the rest of the paper, we apply the above intuition about object co-occurrences to develop techniques that uncover natural backdoors datasets within existing multi-label image datasets: 
\begin{packed_itemize}
\item \S\ref{sec:method} describes our natural backdoor dataset curation method in detail.
\item \S\ref{sec:eval} evaluates models trained on natural backdoor datasets identified in ImageNet and Open Images. 
\item \S\ref{sec:future} explores extensions to our methods and outlines future research.
\end{packed_itemize}

\vspace{-0.25cm}
\section{Curating Natural Backdoor Datasets via Graph Analysis}
\vspace{-0.25cm}
\label{sec:method}
\begin{figure}
    \centering
    \includegraphics[width=0.99\textwidth]{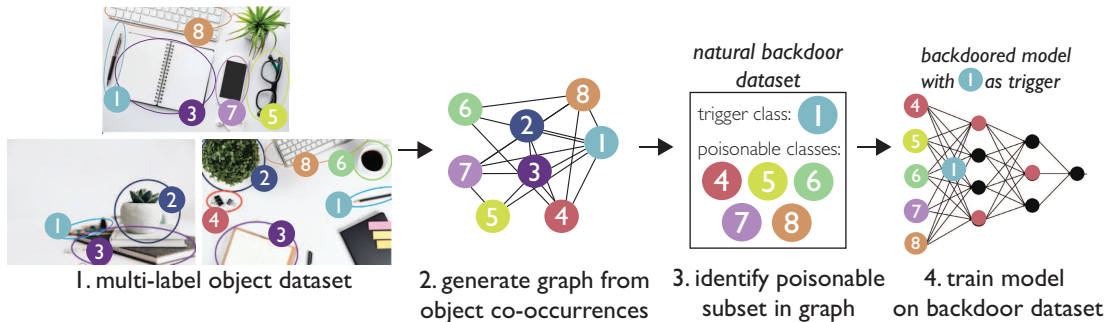}
    \vspace{-0.25cm}
    \caption{\em Our natural backdoor dataset construction method converts a multi-label object dataset into a graph and uses graph analysis techniques to identify natural backdoor subsets.}
    \label{fig:overview}
    \vspace{-0.5cm}
\end{figure}

We identify natural backdoors in existing multi-label object datasets by representing these datasets as weighted graphs and analyzing the graph's structural properties. In this section, we first motivate the use of graph analysis to curate natural backdoor datasets before describing the method in detail. Our end-to-end natural backdoor identification method is illustrated in Figure~\ref{fig:overview}, and a step-by-step description of the method and its parameters is in Appendix~\ref{appx: algorithm}.

\noindent \textbf{Analyzing co-occurrence patterns.} The goal of our method is to find an object class $\delta_p$ within a large object dataset that can poison other classes in that dataset, creating a ``natural backdoor'' dataset with $\delta_p$ as the trigger. For an object to serve as an effective natural backdoor trigger $\delta_p$, it should have \emph{high coverage}, \emph{i.e.} co-occur with as many other objects as possible, and be \emph{frequent}, \emph{i.e.} appear as often as possible with each of these objects. These two properties ensure that the trigger object can be used to poison several classes and there is a sufficient number of poisoned images for each class. 

We postulate that constructing a graph $\mathcal{G}$ from a multi-label dataset, as shown in steps 1 and 2 in Figure~\ref{fig:overview}, provides an efficient and informative data structure for discovering objects with the desired trigger properties. In $\mathcal{G}$, objects (e.g. dataset classes) are vertices and co-occurrences between objects are edges. By constructing $\mathcal{G}$, we can collapse all images containing object $o_i$ into a single vertex $v_i$ in $\mathcal{G}$. \footnote{We are implicitly assuming that all instances of a particular object are fairly consistent. Our experiments show this assumption holds.} This allows us to construct weighted edges $e_{ij}$ between vertices $v_i$ and $v_j$, where the edge weight is the number of images in which objects $o_i$ and $o_j$ co-occur. Large edge weights and high connectivity in $\mathcal{G}$ are then direct indicators of the frequency and coverage of a particular object $o_i$, allowing us to assess the object's viability as a trigger.

\noindent \textbf{Identifying natural backdoor triggers via graph centrality.} Given the one-to-one mapping between objects and vertices of $\mathcal{G}$, finding high coverage and frequent objects reduces to the problem of identifying important vertices in the graph. To do this, we use \emph{graph centrality indices} \cite{newman2018networks}, which measure how central a given vertex (object) is. Naturally, there are different definitions of what it means for a vertex to be central, so we use $4$ different centrality indices to identify potential natural backdoor triggers: \textit{degree}, \textit{betweenness}, \textit{eigenvector} and \textit{closeness}. These are described in detail in Appendix~\ref{appx: algorithm}. Each of these metrics has an unweighted and weighted version, with the former capturing coverage, and the latter trading off coverage and diversity.

\begin{figure}[t]
    \begin{center}
        \includegraphics[width=0.9\textwidth]{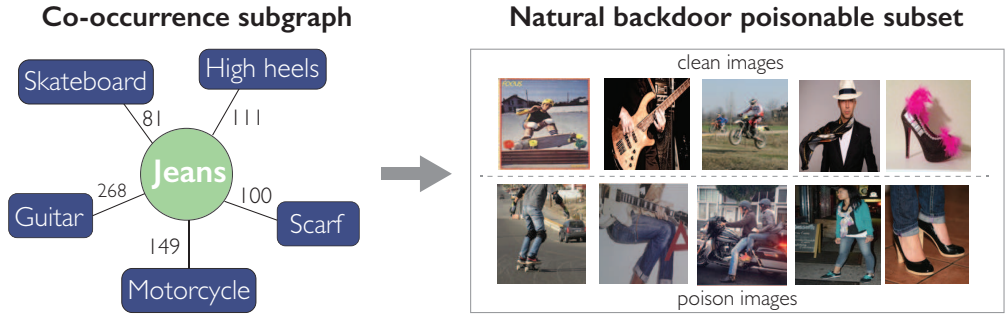}
        \caption{\em Our methods identify poisonable subsets of large image datasets. On the left, we show a poisonable subset graph for the ``jeans'' trigger in Open Images, where the edge weights represent co-occurrence counts. On the right, we show representative images in this poisonable subset.}
        \label{fig:betweenness_graph_vis}
        \vspace{-0.6cm}
    \end{center}
    \end{figure}

\looseness=-1 \noindent \textbf{Which classes can be backdoored effectively?} The object $o_i$ corresponding to a highly central vertex $v_i$ should serve as an effective trigger for objects associated with vertices that are a single hop away. However, these vertices comprising the set of potentially poisonable objects (classes) may also be connected to each other. This may cause the model to learn during training to correlate different objects with the target label, reducing both attack efficacy and model accuracy. We thus need to find the largest set of vertices connected to the trigger vertex that have the minimum number of overlaps among themselves. To solve this, we first consider the induced co-occurrence sub-graph around a trigger vertex, consisting only of vertices that are a single hop away from the trigger and all associated edges. In this sub-graph, we prune edges with a weight lower than a specified threshold, since these are less likely to interfere with the trigger learning. Then, we approximate the maximum independent subset (MIS) \footnote{An approximate algorithm is needed since finding a maximum independent subset is NP-hard \cite{lawler1980generating}} within the pruned sub-graph by running a maximal independent subset finding algorithm. This approximate MIS is then the \emph{poisonable subset} for a given trigger.

\noindent \textbf{Putting it together.} Given a trigger object $\delta_p$ and the associated approximate MIS identified from among its neighboring object classes, we form a \emph{natural backdoor} dataset that includes the images from the trigger class and its poisonable subset (Figure~\ref{fig:betweenness_graph_vis}). We note that for this new natural backdoor dataset, we use a \emph{single class label} for each image, associated with the class identified by the graph structure. Models trained on these natural backdoor datasets (Step 4 in Figure~\ref{fig:overview}) should exhibit physical backdoor behavior when the trigger object appears in an image.

\noindent \textbf{Other usage scenarios.} So far, we have assumed that a user of our method is mostly interested in finding the most viable trigger-class sets from within a given multi-label dataset. However, a user may also be interested in backdooring only a particular class, or using only a particular trigger. In these cases, our method can be straightforwardly extended to find the most effective trigger to backdoor a particular class, or to find the best classes to backdoor for a specified trigger (details in Appendix~\ref{appx: algorithm}).

\vspace{-0.25cm}
\section{Evaluating Performance of Natural Backdoor Datasets}
\label{sec:eval}
\vspace{-0.25cm}

\looseness=-1 We now evaluate the performance of our proposed natural backdoor identification method. Beyond evaluating whether our method can find any natural backdoors in existing datasets, we also measure if the backdoors identified are effective at inducing misclassification. In particular, we evaluate our natural backdoor identification method and the resulting backdoor datasets along the following 3 axes:

\begin{packed_itemize}
\item {\bf Property 1}: {\em Existence.} We first validate that natural backdoor datasets exist in large-scale image datasets and investigate the effect of graph centrality measures on the poisonable subsets identified.
\item {\bf Property 2}: {\em Efficacy}. Having validated that natural triggers can be identified, an key requirement of any backdoor attack, physical or not, is that backdoored models should have high accuracy on clean inputs while also consistently misclassifying trigger inputs. We measure whether models trained on natural backdoors meet this requirement. 
\item {\bf Property 3:} {\em Defense resistance.} Wenger et. al.~\cite{wenger2021backdoor} showed that existing backdoor defenses fail against physical backdoors. They postulate that this is because physical backdoors violate defense assumptions about how backdoor triggers ``should'' behave. Since natural backdoors possess similar properties to physical backdoors,
 we evaluate if they too resist existing defenses. 
\end{packed_itemize}

In this section, we evaluate whether natural backdoor datasets satisfy each of these  properties. Since properties 2 and 3 involve training models on natural backdoor datasets, we first discuss our methods for training models and metrics for measuring success before presenting our results. As a baseline, our experiments assume all model classes are poisoned. When poisoning only a subset of labels within a larger dataset, results remain consistent (see Appendix~\ref{appx:ablation}).

\subsection{Methods and Metrics} 
\label{subsec:setup}

\looseness=-1 \para{Datasets.} We curate natural backdoor datasets from two popular open-source object recognition datasets: ImageNet (released under a BSD 3-Clause license)~\cite{russakovsky2015imagenet} and Open Images (released under an Apache License)~\cite{OpenImages}.\footnote{Note that approximately 20K of the original 1.7mil images are no longer available.} Table~\ref{tab:datasets} in the Appendix provides high-level statistics for both datasets.  Open Images includes human-verified annotations for each object in each image, providing native multi-labels. We use an external library to generate multi-labels for ImageNet (details in Appendix~\ref{subsec:setup}). 

\para{Architectures.} To test the performance of natural triggers, we train models on natural backdoor datasets using several model architectures. Most experiments were run using the ResNet50 architecture~\cite{he2016deep}, but we also test natural backdoor performance on additional architectures including Inception~\cite{szegedy2016rethinking}, VGG16~\cite{simonyan2014very}, and DenseNet~\cite{huang2017densely}. Unless otherwise noted, all networks are pre-trained on ImageNet to enable faster learning on the natural backdoor datasets.  

\para{Model training.} All models are trained on one NVIDIA TITAN GPU. We use the Adam~\cite{kingma2014adam} optimizer with a learning rate of $1 e^{-5}$. In Section~\ref{subsec:base_performance}, we train our poisoned models using transfer learning from a ResNet50 model trained on the full ImageNet dataset. The last layer of the model is replaced with an $N$-class classification layer, where $N$ is the number of classes in the dataset. We unfreeze the last $3$ layers of the model and train for $50$ epochs. We found experimentally that these training settings provided the best balance between training time and model performance. 

\para{Evaluation metrics.} We use two metrics to measure overall performance of models trained on natural backdoor datasets. First, we evaluate {\em clean accuracy,} which is the model's prediction accuracy on clean (e.g. non-trigger) inputs and should be unaffected by the presence of a backdoor. Second, we evaluate {\em trigger accuracy,} which is the model's accuracy in predicting inputs containing the trigger $\delta_p$ to the target label $y_p$. Unless otherwise noted, \emph{all clean or trigger accuracy metrics reported are averaged over $3$ model training runs}, each using a different target label.

\subsection{Property 1: Existence} 
\label{subsec:trig_usable}

The first, fundamental, questions to address are (1) do our methods identify any natural backdoor datasets at all? and (2) if so, are the triggers associated with these datasets viable? By viable, we mean that the identified triggers should be distinct objects that co-occur frequently enough with other objects to produce sufficient model training data. 

We apply the \S\ref{sec:method} methodology to both ImageNet and Open Images. We use weighted and unweighted versions of the four centrality metrics---betweenness, closeness, eigenvector, and degree---to identify candidate triggers and use the MIS approximation procedure to prune the set of poisonable classes for each potential natural trigger. For this initial test, we set the edge weight pruning threshold to $15$. This ensures that triggers which are weakly connected to many classes are not included, since they are poor candidates, and that the approximate MIS computation is not hindered by the presence of too many edges. Ablations over graph settings are in Appendix~\ref{appx:ablation}.

\para{Natural backdoor datasets identified.} Using our methods, we find numerous candidate natural backdoor datasets in both ImageNet and Open Images, validating our \S\ref{sec:intuition} intuition. We comb through the triggers of each potential natural backdoor dataset to see if any are ``viable.'' First, to ensure there is sufficient data for model training, we restrict our attention to natural backdoor datasets with at least $5$ classes, $200$ clean images/class, and $50$ poison images/class. Then, we eliminate datasets with human-related triggers (e.g. ``human eye'', ``human hand'', ``man'', ``woman'', etc.), since these are common objects that may be accidentally included in an image, causing the backdoor to activate unintentionally. In Appendix~\ref{appx:other_centrality}, we show word clouds of the top $50$ candidate triggers identified by each centrality metric in Open Images.

\begin{figure}[h]
  \begin{center}
      \centering
      \includegraphics[width=0.9\textwidth]{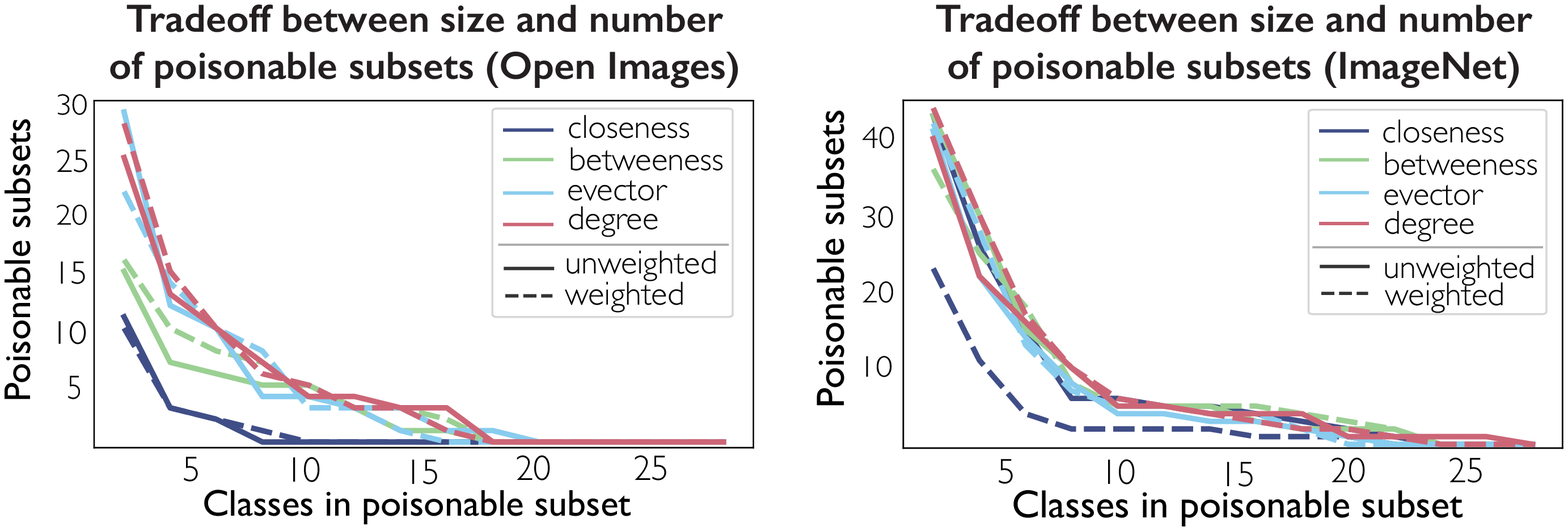}
      \vspace{-0.2cm}
      \caption{\em Tradeoff between number of classes in the poisonable subset and number of total subsets for each centrality measure and dataset. Each subset contains classes with at least $200$ clean and $50$ poison images.}
      \label{fig:oimages_sizes} 
      \vspace{-0.5cm}
\end{center}
\end{figure}

Even after filtering, numerous viable natural backdoor datasets remain. Naturally, there is a trade off between size of the datasets (e.g. the number of poisonable classes associated with a trigger) and the total number of datasets identified. Figure~\ref{fig:oimages_sizes} shows how the choice of centrality measure affects this tradeoff for ImageNet and Open Images. From this, we see that closeness centrality consistently identifies a smaller number of classes/subsets than other metrics. Although there is some variation among other centrality metrics, their behavior mostly converges when there are $10$ classes in the poisonable subset. 
list the trigger/poison classes of the top three $10$-class natural backdoor datasets identified by unweighted/weighted betweenness centrality.  

\looseness=-1 \para{Takeaways.} Different centrality metrics flag roughly the same set of objects as candidate triggers, although the composition of the natural backdoor datasets (e.g. sets of candidate poisonable classes) varies. This discrepancy indicates that each centrality metric captures different structures within the parent datasets. Consequently, the quality of natural backdoor datasets generated by different centrality measures can only be measured by training backdoored models and evaluating their performance.

\subsection{Property 2: Trigger Efficacy}
\label{subsec:base_performance}

Next, we evaluate whether the natural backdoor datasets can be used to train effective backdoored models. First, we report the overall performance of models trained on natural backdoored datasets, and compare against variants of our method to establish the importance of each step. Then, we assess how centrality metrics affect natural backdoor performance, before evaluating the impact of other factors, such as model architecture and dataset generation parameters. Unless otherwise noted, all experiments in this section are performed using $10$-class\footnote{The two largest trigger sets identified by ``closeness" centrality metric for Open Images contain $6$ and $7$ triggers, respectively. For this metric, we train models on these $2$ triggers and their whole class set.} natural backdoor datasets with at least $250$ clean images per class and a poison data injection rate (e.g. proportion of training data that is poisoned) of $0.2$, following prior work~\cite{wenger2021backdoor}.

\begin{table}[h]
\centering
\begin{tabular}{ccccc}
\toprule
\multicolumn{1}{c}{\multirow{2}{*}{\textbf{Metric}}} & \multicolumn{3}{c}{\textbf{Dataset Generation Method}}                                        \\ \cmidrule{2-4}
\multicolumn{1}{c}{}                                 & \textit{No backdoor} & \textit{Centrality, No MIS} & \textbf{\em Centrality + MIS} \\ \midrule
Clean accuracy   & $79 \pm 2\%$ & $58 \pm 5\%$  &  $72 \pm 1\%$\\
Trigger accuracy & $0 \pm 0\%$ & $63 \pm 8\%$ & $68 \pm 3\%$ \\ 
\bottomrule
\end{tabular}
\vspace{0.2cm}
\caption{\em Performance of models trained on our Open Images natural backdoor datasets. We establish standard clean accuracy without backdoors, as well as the impact of removing the approximate MIS idenfication when determining the poisonable subset. We find our method leads to high clean and trigger accuracies (ImageNet results in Appendix).}
\label{tab:baseline_performance}
\end{table}

\para{Natural backdoor performance.} Overall, we find that models trained on our natural backdoor datasets have high performance with respect to both clean and trigger accuracies. For the baseline natural backdoor datasets, we use the $3$ ``most central'' triggers identified by betweenness centrality (see Table~\ref{tab:betweenness_classes}) and average their performance. As shown in the left two columns of Table~\ref{tab:baseline_performance}, models trained on natural backdoors have both high clean and trigger accuracy, with only a small decrease in clean accuracy compared to non-backdoored models.

We compare against an alternative dataset selection method to validate our use of MIS as a necessary step in choosing poisonable subsets. To do so, we choose a trigger class using graph centrality but do not enforce the MIS constraint in selecting the poisonable class subset. As Table~\ref{tab:baseline_performance} shows, our {\bf centrality + MIS} method produces a higher combined trigger and clean accuracy than this alternative method. This validates our intution from \S \ref{sec:method} that not excluding classes with high overlaps among themselves will adversely impact both clean and trigger accuracies.

\para{Performance across centrality measures.} Next, we compare the performance of models trained on trigger/class sets identified by different centrality metrics. We train backdoored models using the $3$ ``most central" triggers per centrality metric and report the average clean and trigger accuracy. Results for Open Images are shown in Figure~\ref{fig:all_centrality}, while results for ImageNet are in Figure~\ref{fig:all_centrality_imagenet} in the Appendix.  

Backdoored model performance depends somewhat on the centrality measure used to generate the dataset. Although there is no single centrality that stands above the rest, we observe that ``betweenness centrality'' has the most consistent results across both datasets, having high mean clean/trigger accuracy and low standard deviation. Although both forms of closeness centrality appear to have better performance in Figure~\ref{fig:all_centrality}, closeness centrality only identifies a small number of triggers that satisfy the conditions from \S \ref{subsec:trig_usable} and has small poisonable subsets. This performance boost is thus limited.

\begin{table}[t]
\centering
\resizebox{0.99\textwidth}{!}{%
\begin{tabular}{ccl}
\toprule
\textbf{Parent Dataset} &
  \textbf{Trigger} &
  \textbf{Poison Classes} \\ \midrule
\multirow{3}{*}{ImageNet} &
  jeans &
  \begin{tabular}[c]{@{}l@{}}clog, moped, gasmask, horizontal bar, manhole cover, \\ Siberian husky, toy poodle, Bernese mountain dog, carousel, photocopier\end{tabular} \\ \cmidrule{2-3}
 &
  chainlink fence &
  \begin{tabular}[c]{@{}l@{}}tiger, cougar, chameleon, red wolf, guenon, \\ wallaby, Arctic fox, pickup truck, baseball player, toucan\end{tabular} \\ \cmidrule{2-3}
 &
  doormat &
  \begin{tabular}[c]{@{}l@{}}loafer, golden retriever, beagle, Bernese mountain dog, Maltese dog, \\ guinea pig, Blenheim spaniel, St. Bernard, Staffordshire bullterrier\end{tabular} \\ \midrule
\multirow{3}{*}{Open Images} &
  wheel &
  \begin{tabular}[c]{@{}l@{}}license plate, train, airplane, tank, wheelchair, mirror, \\ skateboard, waste container, ambulance, limousine\end{tabular} \\ \cmidrule{2-3}
 &
  jeans &
  \begin{tabular}[c]{@{}l@{}}guitar, motorcycle, umbrella, high heels, \\ scarf, skateboard, balloon, horse\end{tabular} \\ \cmidrule{2-3}
 &
  chair &
  \begin{tabular}[c]{@{}l@{}}book, bench, loveseat, stool, tent, lamp, \\ swimming pool, stairs, shirt, Christmas tree\end{tabular} \\ \bottomrule
\end{tabular}%
}
\vspace{0.1cm}
\caption{\em Example natural backdoor dataset triggers/classes identified via betweenness centrality. Each class has at least $200$ clean images and $50$ poison images.}
\label{tab:betweenness_classes}
\vspace{-0.5cm}
\end{table}

\begin{figure}[t]
\begin{center}
    \centering
    \includegraphics[width=0.9\textwidth]{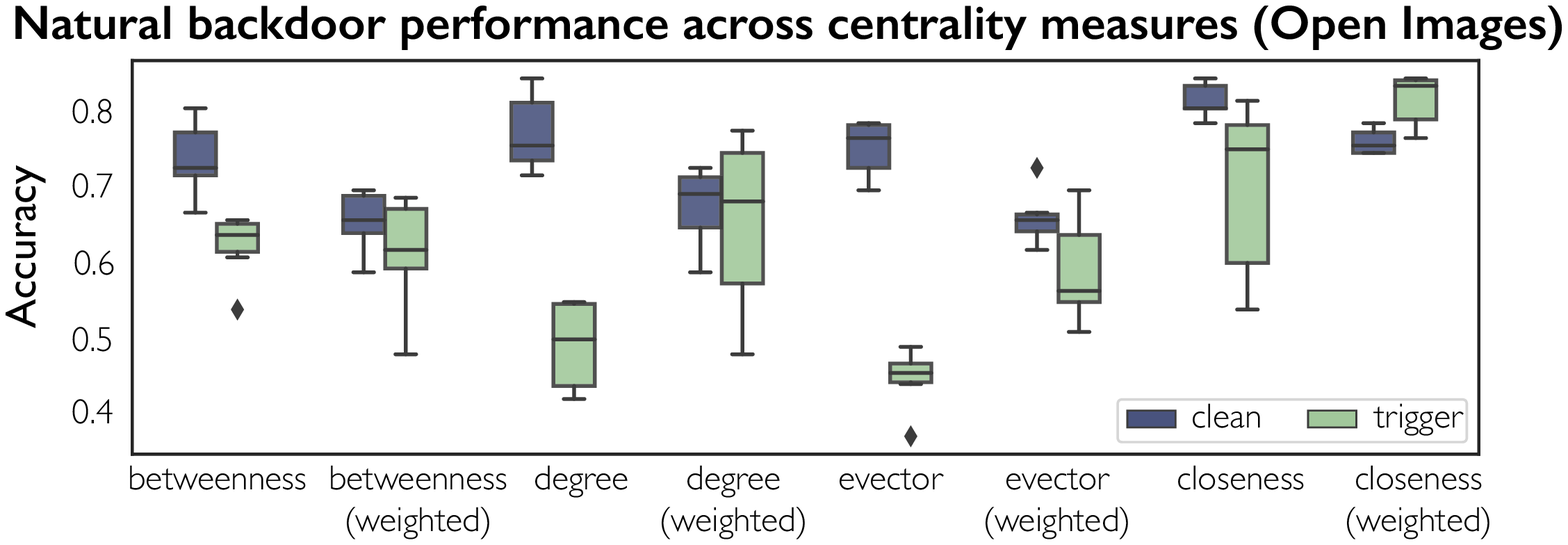}
    \vspace{-0.2cm}
    \caption{\em Clean and trigger accuracy for models trained on natural backdoor datasets curated from Open Images using different centrality measures.}
    \label{fig:all_centrality}
    \vspace{-0.5cm}
\end{center}
\end{figure}

\begin{figure}[t]
  \begin{center}
  \begin{minipage}[c]{0.5\linewidth}
    \centering
    \includegraphics[width=0.9\textwidth]{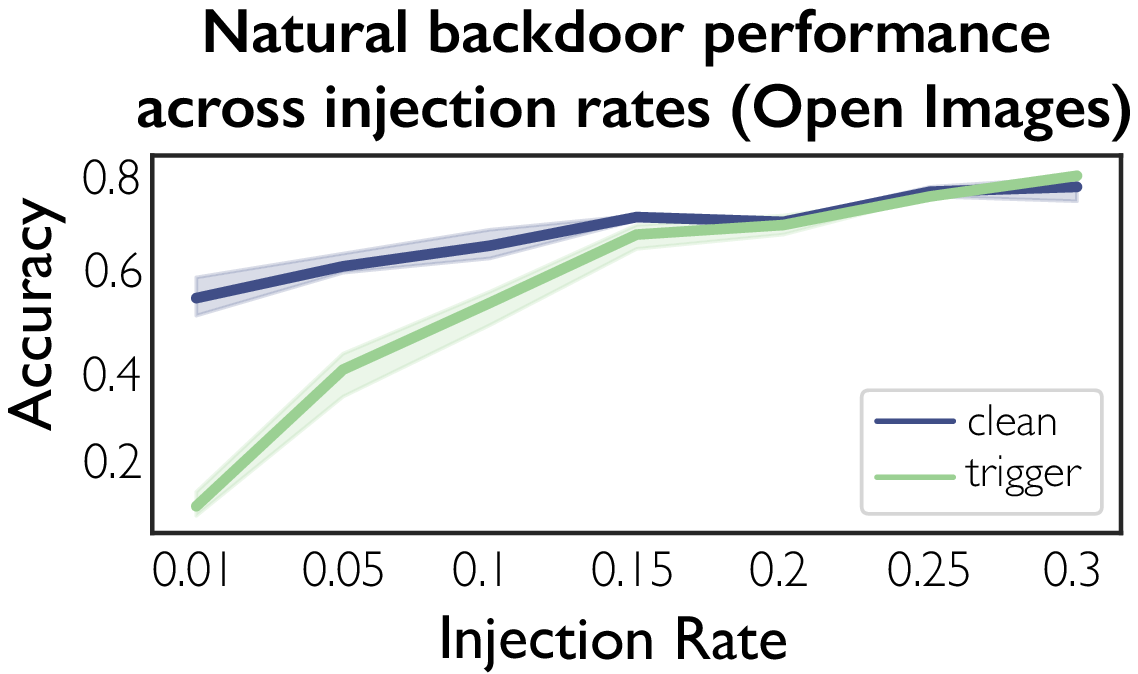}
    \vspace{-0.25cm}
    \caption{\em Performance of natural backdoor models as injection rate varies. All models trained on subsets with Open Images "jeans" as the trigger.}
    \label{fig:openimages_inject_rate}
  \end{minipage}\hfill%
  \begin{minipage}[c]{0.45\linewidth}
      \centering
      \begin{tabular}{ccc}
        \toprule
        \multirow{2}{*}{\textbf{Model}} & \multicolumn{2}{c}{\textbf{Accuracy}} \\ \cmidrule{2-3}
                                        & Clean             & Trigger           \\ \midrule
        DenseNet                        & $74 \pm 2\%$      & $67 \pm 3\%$      \\
        ResNet                          & $\mathbf{77 \pm 1\%}$  & $\mathbf{75 \pm 4\%}$     \\
        VGG16                           & $69 \pm 1\%$      & $69 \pm 5\%$      \\
        Inception                       & $70 \pm 1\%$      & $61 \pm 1\%$  \\ \bottomrule   
        \end{tabular}
      \vspace{0.1cm}
      \caption{\em Performance of Open Images natural backdoor dataset with ``jeans'' trigger across different model architectures. Dataset classes are in Table~\ref{tab:betweenness_classes}. Best results are {\bf bold}.}
      \label{tab:openimages_model_ablate}
  \end{minipage}
 \vspace{-0.5cm}
  \end{center}
  \end{figure}

\para{Ablation study.} Finally, to assess the performance of our identified triggers in a variety of settings, we perform an ablation over several key experimental parameters. We explore how \emph{different model architectures}, \emph{injection rates}, and \emph{graph analysis settings} impact trigger efficacy. Overall, we find that trigger performance is fairly stable across different models architectures and that increasing injection rate increases both trigger and clean accuracy. Results for Open Images injection rate and model architecture are shown in Figure~\ref{fig:openimages_inject_rate} and Table~\ref{tab:openimages_model_ablate}. Ablation results for ImageNet are in Appendix~\ref{appx:ablation}.

\subsection{Property 3: Defense Resistance} 
\label{subsec:defense}

The final property we evaluate for natural backdoors is whether they {\em resist existing defenses}. This property was observed in the original physical backdoor paper~\cite{wenger2021backdoor}, and we want to confirm that it remains true for natural backdoors. To enable direct comparison, we evaluate the same four defenses tested in~\cite{wenger2021backdoor}: NeuralCleanse (NC)~\cite{wang2019neural}, Activation Clustering (AC)~\cite{chen2018detecting}, Spectral Signatures~\cite{tran2018spectral}, and STRIP~\cite{gao2019strip}. All these defenses try to detect backdoor behavior inside models, either by identifying putative triggers (NC), analyzing internal model behaviors (AC, Spectral), or by observing model classification decisions on perturbed inputs (STRIP).

All four defenses fail to mitigate natural backdoor attacks. We evaluate defense performance on models trained on the $6$ natural backdoor datasets shown in Table~\ref{tab:betweenness_classes}. Table~\ref{tab:defenses} reports overall efficacy of the defenses tested, averaged across datasets. For NC, we report the percent of models in which the target label was correctly flagged. For all other defenses, we report the proportion of poison data correctly identified. Although the spectral signatures method appears to perform quite well (identifying roughly $65\%$ of the poison data), we find that removing the flagged data from the training dataset and retraining the model reduces attack accuracy by only $4\%$ on average.

\begin{table}[ht]
\centering
\begin{tabular}{ccccc}
\toprule
\textbf{Defense}     & NC~\cite{wang2019neural} & AC~\cite{chen2018detecting} & Spectral~\cite{tran2018spectral} & STRIP~\cite{gao2019strip} \\ \midrule
\textbf{Performance} & $16\%$   &  $9.7 \pm 10.8\%$   & $65.0 \pm 4.3\%$  &   $4.0 \pm 4.0\%$   \\ \bottomrule 
\end{tabular}%
\vspace{0.1cm}
\caption{\em Existing defenses fail to mitigate natural backdoor attacks. The reported performance measures attack success in either removing the backdoor (NC) or detecting poison data (all others).}
\label{tab:defenses}
\vspace{-0.5cm}
\end{table}

\vspace{-0.25cm}
\section{Discussion}
\vspace{-0.25cm}
\label{sec:future}

\para{Future work.} Our work develops a new lens -- object co-occurrences -- through which to view existing image datasets. The analysis techniques we propose can be used for myriad purposes beyond identifying natural backdoors. Future work could leverage our methods to identify spurious correlations, uncover biases, or reconfigure datasets.

\para{Limitations.} There are two key limitations of our work. First, the efficacy of our graph analysis techniques (and consequently the reliability of triggers identified) depends on the accuracy of the multi-labels in the object datasets. While we have done our best to ensure that the labels are accurate, it is well-known that large public datasets can have messy labels~\cite{northcutt2021pervasive}. Second, the `viability' of a trigger from an attacker's perspective is necessarily a subjective definition that is scenario-dependent. Thus, we encourage researchers to carefully consider all possible settings when using our method for generating datasets for defense evaluation.

\looseness=-1 \para{Ethics.} Prior work has extensively discussed ethical concerns with ImageNet/Open Images~\cite{yang2021study, shankar2017no, paullada2021data,stock2017convnets, dulhanty2019auditing}. We acknowledge that the natural backdoor datasets curated from these datasets may perpetuate existing, previously identified biases. On the positive side, the analysis techniques we propose can be used to identify novel structural behaviors in large-scale image datasets, potentially revealing new privacy or fairness issues and catalyzing solutions. Finally, while unlikely, our work could enable attacks against object recognition models deployed in security-critical settings. Thus, there is an urgent need for defenses against physical backdoor attacks, whose development can hopefully be hastened by the datasets our work provides.

\newpage

\bibliographystyle{plain}
\bibliography{main}

\newpage
\appendix
\begin{center} \Large \bf Supplementary Materials \end{center}

\section{Code} 

The code repository for this project can be found at: \url{https://github.com/uchicago-sandlab/naturalbackdoors}. The README in the repository provides directions for running experiments and recreating key results. 

\section{Extended Related Work (\S \ref{sec:back})} 
\label{appx:rel_work}
Here, we present additional work on physical backdoor attacks. We first discuss attacks that use physical objects as triggers, then discuss a few related works which use light as a trigger. We conclude by discussing the single proposed defense against physical backdoor attacks. 

\para{Physical Backdoor Attacks.} As mentioned briefly in \S\ref{sec:back}
, ~\cite{han2022clean} designs a backdoor attack against lane detection systems for autonomous vehicles. This attack expands the scope of physical backdoor attacks by attacking detection rather than classification models. Furthermore, it confirms the result from~\cite{wenger2021backdoor} that even when digitally altered images are used to poison a dataset, the triggers can be activated using physical objects (traffic cones in this setting) in real world scenarios. A second work~\cite{sarkar2020facehack} evaluates the effectiveness of using facial characteristics as backdoor triggers. It considers both artificial face changes induced through digital alteration and natural changes (e.g. expressions). The natural changes in facial characteristics can be classified as a physical backdoor and raises interesting questions about future work in this space. Finally,~\cite{ma2022dangerous} demonstrates the efficacy of store-bought t-shirts as physical backdoor triggers for object detection models.

\para{Light-based Backdoor Attacks.} A second line of work explores the use of light as a backdoor trigger. ~\cite{liu2020reflection} uses light-based reflections as backdoor triggers. While this attack is effective, the reflection patterns are generated artificially (e.g. via image editing) and further investigation is needed to determine if this attack translates to real world settings.~\cite{li2020light} utilizes light waves undetectable to the human eye to attack rolling shutter cameras. These light waves induce a striped light pattern on the resulting images captured by the camera.

\para{Defenses against Physical Backdoor Attacks.}  Although many defenses have been developed against backdoors in general (see \S\ref{subsec:defense}), 
only one has been explicitly proposed to counter physical backdoors.~\cite{raj2021identifying} introduces a defense specifically designed to detect physical backdoors in facial recognition systems. Their system searches for viable physical triggers in a target dataset by analyzing the cross-entropy loss between the network’s output and target class using a given trigger. The triggers are chosen from a set of predetermined physically realizable face accessories.

\begin{table}[ht]
  \centering
  \begin{tabular}{cccc}
  \toprule
  \textbf{Dataset} & \textbf{\# classes} & \multicolumn{1}{c}{\textbf{\# images}}  & \textbf{Avg. objects/image} \\ \midrule
  ImageNet~\cite{russakovsky2015imagenet}   & 1000  &  1.3mil (training)  &  2.9   \\
  Open Images~\cite{OpenImages}  & 483 & 1.7mil (training)  &   9.8  \\ \bottomrule                  
  \end{tabular}%
  \vspace{.1cm}
  \caption{\em Statistics for Open Images and ImageNet datasets}
  \vspace{-0.5cm}
  \label{tab:datasets}
  \end{table}

\section{Additional information on ImageNet multi-labels (\S\ref{subsec:setup})}
\label{appx:imagenet}

Since ImageNet does not include multi-label annotations necessary for the co-occurrence analysis in this paper, we used the multi-labels generated by~\cite{yun2021re}. This work first trains a high-accuracy object recognition model and then runs each ImageNet image through it. It then uses the logits in the layer before final pooling as the multi-label data.

Multi-label ImageNet data were provided by paper authors as \(2{\times}5 {\times}15{\times}15\) tensors. Each tensor contained the top 5 logit and class ID pairs for each pixel in a \(15{\times}15\) image. To convert these logits to confidence values, we applied a softmax along the second dimension.

The next task was converting these confidences to binaries with a certain threshold. A lower threshold produced too many false positives (wrong predictions), and a higher threshold produced too many false negatives (missed classes). Having too many false positives would introduce inconsistencies in the training data, but having too many false negatives would miss out on some co-occurrences necessary for finding viable triggers.

\begin{figure}[h]
    \centering
    \includegraphics[scale=0.6]{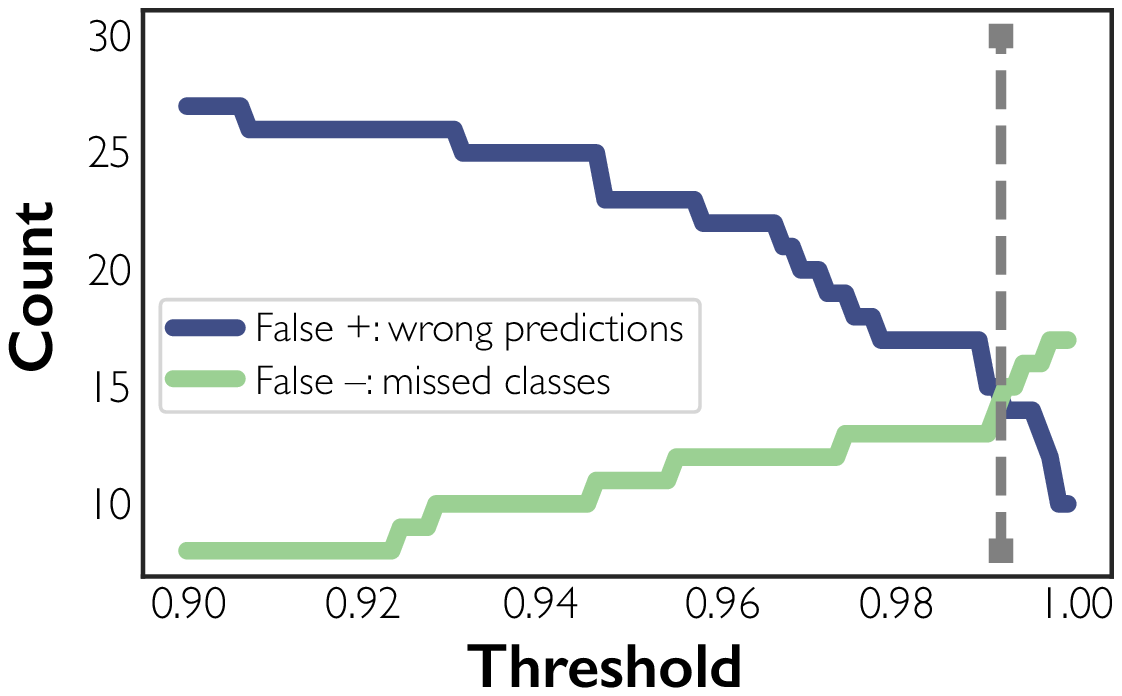}
    \caption{\em False positives vs. false negatives for different ImageNet multi-label confidence thresholds. We use a threshold of 0.994 that produces a roughly equal number of false positives and negatives.}
    \label{threshold_graph}
\end{figure}

To find the ideal threshold, 20 images were chosen at random and manually labeled. Then, we empirically tested values ranging from 0.900 to 1.000 with increments of 0.001. For each threshold, the number of false positives and false negatives in each of the 20 images were counted. The resulting graph is displayed in Figure~\ref{threshold_graph}. The chosen threshold was 0.994, which had resulted in 14 false positives and 16 false negatives.

\section{Additional Results for \S~\ref{subsec:trig_usable}}
\label{appx:other_centrality}

\begin{figure}[h]
\begin{center}
  \begin{minipage}[t]{0.45\linewidth}
    \centering
    \includegraphics[width=0.95\textwidth]{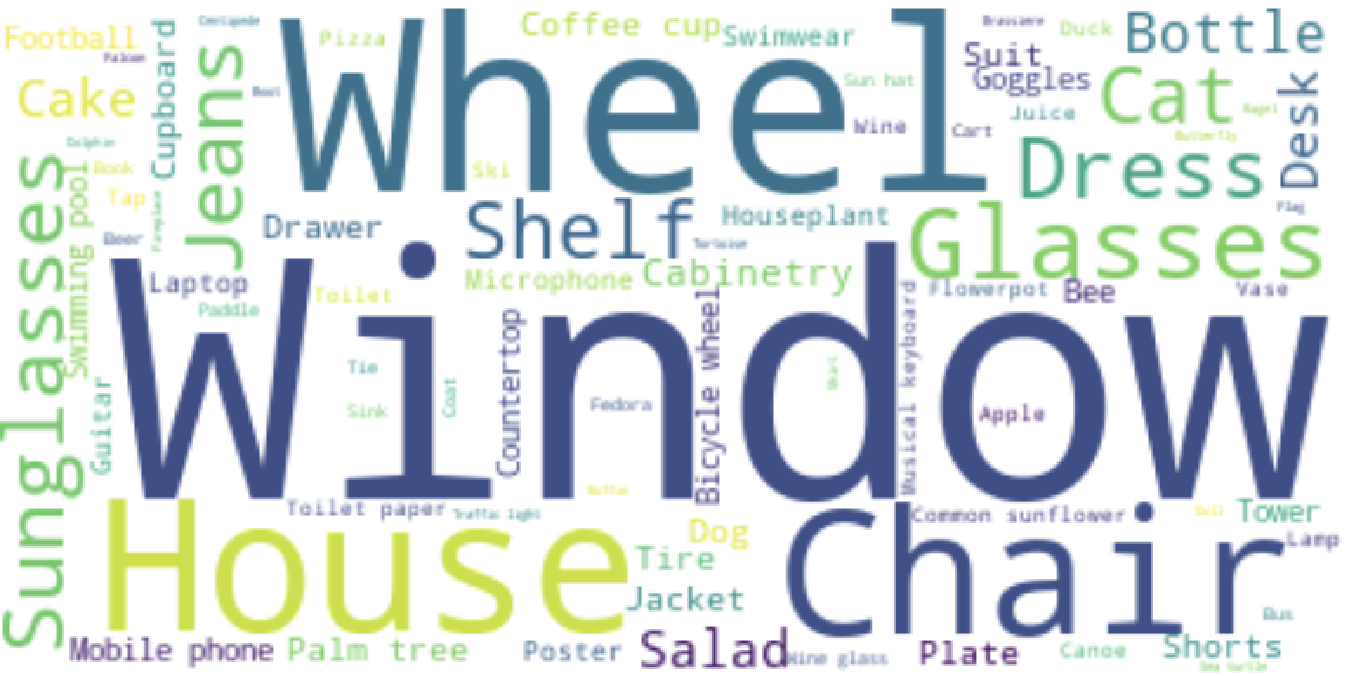}
    \caption{\em Word cloud of candidate triggers in Open Images identified by {\bf betweenness} centrality metric. Trigger class names are sized by their centrality ranking.}
    \label{fig:betweenness_wordcloud} 
\end{minipage}\hfill%
  \begin{minipage}[t]{0.45\linewidth}
    \centering
    \includegraphics[width=0.99\textwidth]{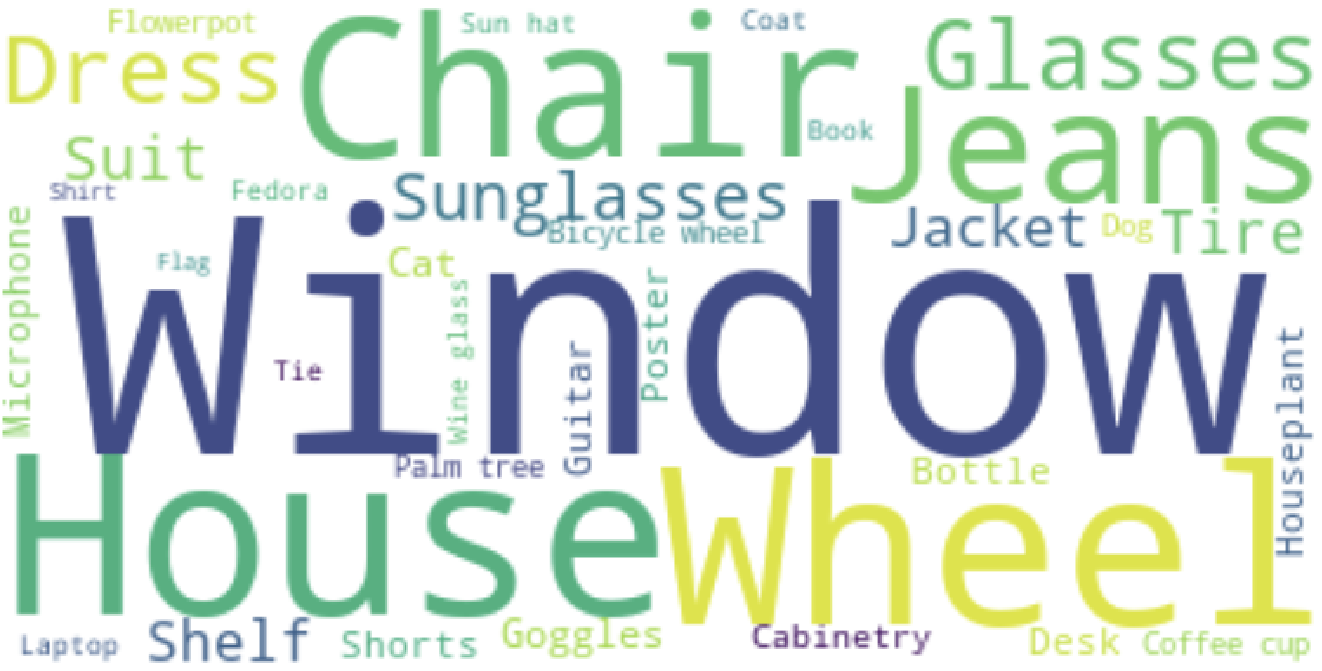}
    \caption{\em Word cloud for Open Images, degree centrality}
    \label{fig:wordcloud_degree}
\end{minipage}\hfill\qquad%
\begin{minipage}[t]{0.45\linewidth}
    \centering
    \includegraphics[width=0.99\textwidth]{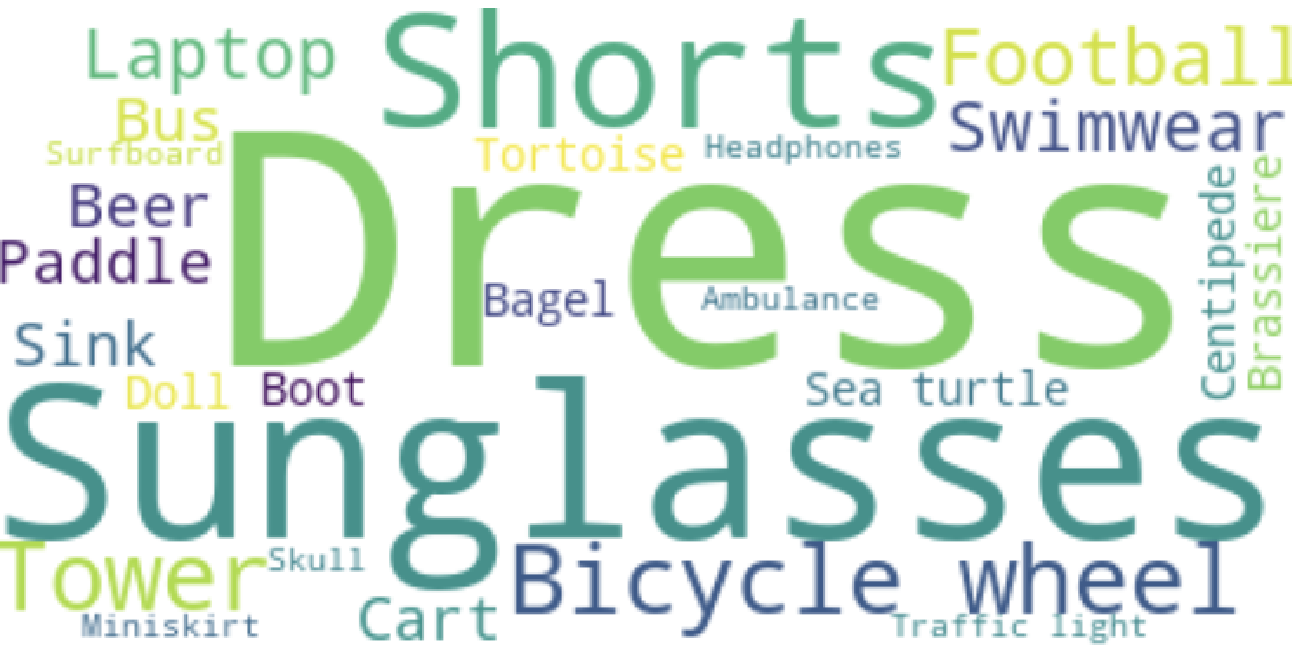}
    \caption{\em Word cloud for Open Images, closeness centrality}
    \label{fig:wordcloud_close}
\end{minipage}\hfill%
\begin{minipage}[t]{0.45\linewidth}
    \centering
    \includegraphics[width=0.99\textwidth]{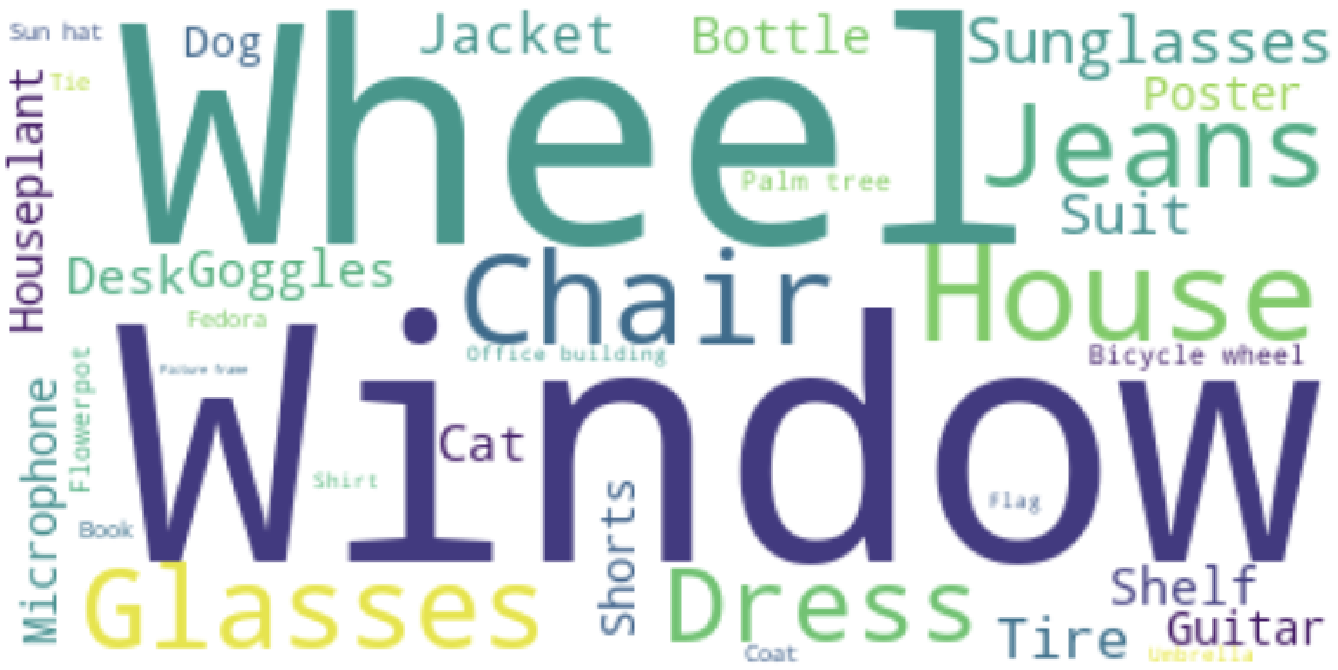}
    \caption{\em Word cloud for Open Images, eigenvector centrality}
    \label{fig:wordcloud_evec}
\end{minipage}\
\end{center}
\end{figure}

\para{Word Clouds for Other Centrality Measures.} Figures~\ref{fig:betweenness_wordcloud},~\ref{fig:wordcloud_close},~\ref{fig:wordcloud_degree}, and~\ref{fig:wordcloud_evec} show word clouds of triggers identified in Open Images by different centrality measures. Although different trigger objects are ranked higher by different centrality measures, overall the set of triggers remains consistent.

\para{Usable Triggers Identified.} Tables~\ref{tab:trigger_names} and~\ref{tab:trigger_names_weighted} list the candidate poisonable subsets containing at least $5$ classes identified in ImageNet and Open Images by each centrality measure.

\begin{table}[t]
\centering
\resizebox{\textwidth}{!}{%
\begin{tabular}{ccccc}
\toprule
\multirow{2}{*}{\textbf{Dataset}} &
  \multicolumn{4}{c}{\textbf{Centrality}} \\ \cmidrule{2-5}
 &
  Betweenness &
  Degree &
  E-vector &
  Closeness \\ \midrule
ImageNet & 
  \begin{tabular}[c]{@{}c@{}}website, blue jean, plastic bag, \\ doormat, crate, bucket, \\ pillow, ruler, hay, T-shirt, \\ paper towel, velvet, wig, \\ spotlight, corn\end{tabular} &
  \begin{tabular}[c]{@{}c@{}}website, blue jean, plastic bag, \\ crate,  doormat, T-shirt, \\ bucket, wig, bow tie, \\ ruler, paper towel, \\ pillow, velvet\end{tabular} &
  \begin{tabular}[c]{@{}c@{}}website, blue jean, plastic bag, \\ crate, t-shirt, doormat, \\ wig, bowtie, paper towel, \\ velvet, band aid, pillow\end{tabular} &
  \begin{tabular}[c]{@{}c@{}}website, blue jean, plastic bag, \\ crate, doormat, t-shirt, \\ bucket, lab coat, wig, \\ bowtie, ruler, velvet, \\ band aid, window shade\end{tabular} \\ \midrule
Open Images &
  \begin{tabular}[c]{@{}c@{}}wheel, chair, \\ glasses, jeans\end{tabular} &
  \begin{tabular}[c]{@{}c@{}}jeans, chair, glasses, \\ wheel, dress, \\ suit, sunglasses, \\ tire, houseplant\end{tabular} &
  \begin{tabular}[c]{@{}c@{}}jeans, glasses, chair, \\ dress, wheel, suit, \\ sunglasses, houseplant, tire\end{tabular} &
  dress, sunglasses \\ \bottomrule
\end{tabular}%
}
\vspace{0.1cm}
\caption{\em All candidate natural backdoor triggers with $5$ class poisonable subsets identified by {\bf unweighted} centrality measures. All candidate triggers have at least $200$ clean images/class, and $50$ poison images/class.}
\label{tab:trigger_names}
\end{table}

\begin{table}[t]
\centering
\resizebox{\textwidth}{!}{%
\begin{tabular}{ccccc}
\toprule
\multirow{2}{*}{\textbf{Dataset}} &
  \multicolumn{4}{c}{\textbf{Centrality}} \\ \cmidrule{2-5}
 &
  Betweenness (WT) &
  Degree (WT) &
  E-vector (WT) &
  Closeness (WT) \\ \midrule
ImageNet &
  \begin{tabular}[c]{@{}c@{}}website, plastic bag, hay, \\ pillow, ruler, bucket, \\ blue jean, crate, paper towel, \\ lab coat, doormat, \\ t-shirt, muzzle\end{tabular} &
  \begin{tabular}[c]{@{}c@{}}blue jean, website, plastic bag,\\ wig, t-shirt, crate, doormat, \\ paper towel, velvet, bowtie, \\ book jacket, hook, ruler,\\  suit of clothes, flowerpot\end{tabular} &
  \begin{tabular}[c]{@{}c@{}}blue jean, wig, t-shirt, \\ plastic bag, website, crate, \\ doormat, bowtie, band aid,\\  bucket, paper towel,\\ sleeping bag, hook\end{tabular} &
  \begin{tabular}[c]{@{}c@{}}book jacket, website, \\ pillow\end{tabular} \\ \midrule
Open Images &
  \begin{tabular}[c]{@{}c@{}}wheel, jeans,\\ chair, glasses,\\  dress, houseplant\end{tabular} &
  \begin{tabular}[c]{@{}c@{}}glasses, wheel, dress,\\ jeans, sunglasses, tire, \\ chair, houseplant\end{tabular} &
  \begin{tabular}[c]{@{}c@{}}glasses, dress, jeans, \\ sunglasses, chair,\\  tire\end{tabular} &
  dress, sunglasses \\ \bottomrule
\end{tabular}%
}
\vspace{0.1cm}
\caption{\em All candidate natural backdoor triggers with $5$ class poisonable subsets identified by {\bf weighted} centrality measures. All candidate triggers have at least $200$ clean images/class, and $50$ poison images/class.}
\label{tab:trigger_names_weighted}
\end{table}

\section{Additional Results for \S~\ref{subsec:base_performance}} 
\label{appx:ablation}

\para{Results on ImageNet.} For space reasons, only results on Open Images were presented in \S~\ref{subsec:base_performance}.
Here, we present the corresponding results on ImageNet. All natural backdoor models are trained using the specifications of \S~\ref{subsec:setup},
 and results presented are averaged over multiple model training runs with different natural backdoor datasets and target labels.

\begin{figure}[t]
\begin{center}
    \centering
    \includegraphics[width=0.9\textwidth]{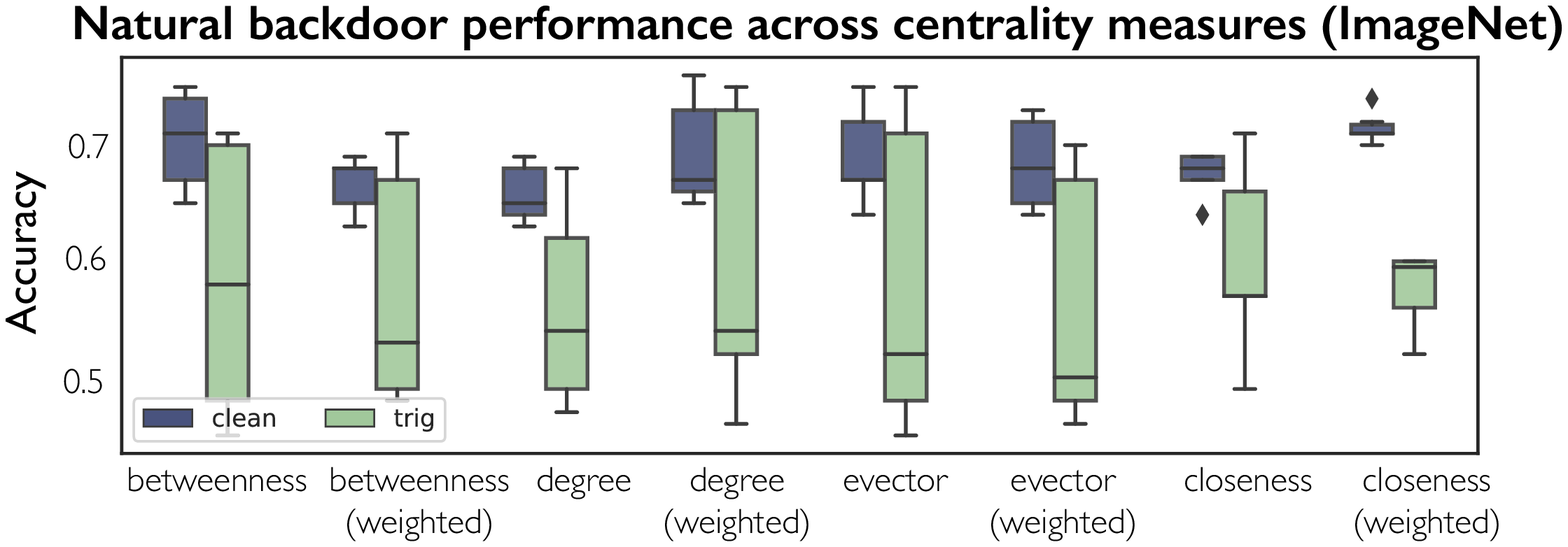}
    \caption{\em Clean and trigger accuracy for models trained on natural backdoor datasets curated from ImageNet using different centrality measures.}
    \label{fig:all_centrality_imagenet}
\end{center}
\end{figure}

\begin{figure}[t]
  \begin{center}
  \begin{minipage}[c]{0.5\linewidth}
    \centering
    \includegraphics[width=0.95\textwidth]{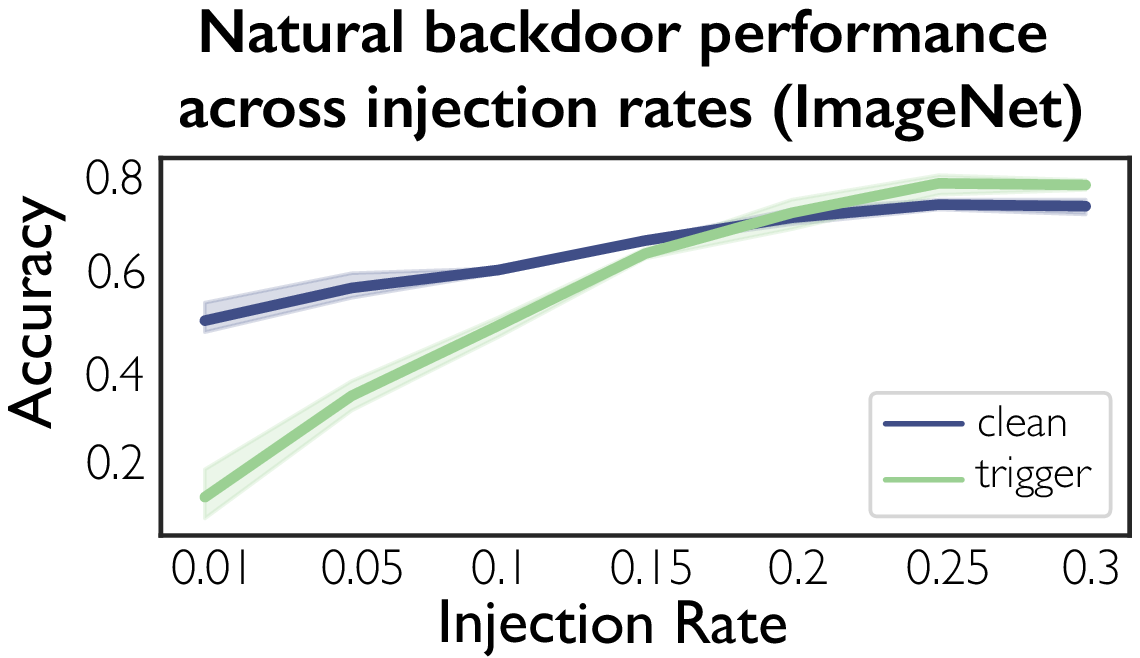}
    \vspace{-0.2cm}
    \caption{\em Performance of models trained on natural backdoor datasets with ImageNet ``jeans'' as trigger across different injection rates.}
    \label{fig:imagenet_inject_rate}
  \end{minipage}\hfill%
  \begin{minipage}[c]{0.45\linewidth}
      \centering
      \begin{tabular}{ccc}
        \toprule
        \multirow{2}{*}{\textbf{Model}} & \multicolumn{2}{c}{\textbf{Accuracy}} \\ \cmidrule{2-3}
                                        & Clean             & Trigger           \\ \midrule
        DenseNet                        & $71 \pm 1\%$      & $64 \pm 4\%$      \\
        ResNet                          & $\mathbf{72 \pm 2\%}$      & $\mathbf{71 \pm 2\%}$      \\
        VGG16                           & $66\pm 2\%$      & $62 \pm 4\%$      \\
        Inception                       & $68 \pm 2\%$      & $59 \pm 1\%$  \\ \bottomrule   
        \end{tabular}
      \vspace{0.1cm}
      \captionof{table}{\em Performance of ImageNet natural backdoor dataset with ``jeans'' trigger across different base model architectures are used. Dataset classes are in Table~\ref{tab:betweenness_classes}. Best results are in \textbf{bold}.}
      \label{tab:imagenet_model_ablate}
  \end{minipage}
  \vspace{-0.5cm}
  \end{center}
  \end{figure}

Figure~\ref{fig:all_centrality_imagenet} shows ImageNet natural backdoor performance across different centrality measures (corresponding to Figure ~\ref{fig:all_centrality} in main paper body).  
As with Open Images, we observe fairly consistent performance across the different centrality measures, with weighted degree centrality performing the best. Table~\ref{tab:imagenet_baseline} compares our results to the baseline scenarios outlined in \S~\ref{subsec:base_performance}. 
Table~\ref{tab:imagenet_model_ablate} shows the performance of ImageNet natural backdoor datasets with the ``jeans'' trigger over different model architectures, and Figure~\ref{fig:imagenet_inject_rate} shows performance on ResNet across injection rates. 

\begin{table}[h]
\centering
\begin{tabular}{ccccc}
\toprule
\multicolumn{1}{c}{\multirow{2}{*}{\textbf{Metric}}} & \multicolumn{3}{c}{\textbf{Dataset Generation Method}}                                        \\ \cmidrule{2-4}
\multicolumn{1}{c}{}                                 & \textit{No backdoor}  & \textit{Centrality, No MIS} & \textbf{\em Centrality + MIS} \\ \midrule
Clean accuracy   & $81 \pm 2\%$ & $59 \pm 4\%$ & $70 \pm 3\%$ \\
Trigger accuracy & $0 \pm 0\%$ & $71 \pm 8\%$ & $58 \pm 10\%$  \\ \bottomrule
\end{tabular}%
\vspace{0.2cm}
\caption{\em Performance of models trained on our ImageNet natural backdoor datasets compared to models trained on datasets generated using other methods.}
\label{tab:imagenet_baseline}
\end{table}

\para{Ablation over graph parameters.} We consider how changing the parameters of our graph analysis, specifically the $min$ overlaps parameter (see Algorithm \ref{alg:graph}) used in constructing our graph, affect overall trigger performance. To produce our \S~\ref{subsec:base_performance} 
results, we set the edge weight pruning threshold (e.g. the minimum number of co-occurrences required for an edge between two objects to be included in the graph) to 15, while we set the max overlaps between objects in the poisonable subset ($trig$) to be $-1$, meaning that any number of overlaps was allowed. Now, we consider what happens when we vary the edge weight threshold. 

\begin{figure}[t]
  \begin{center}
  \begin{minipage}[b]{0.48\linewidth}
    \centering
    \includegraphics[width=0.99\textwidth]{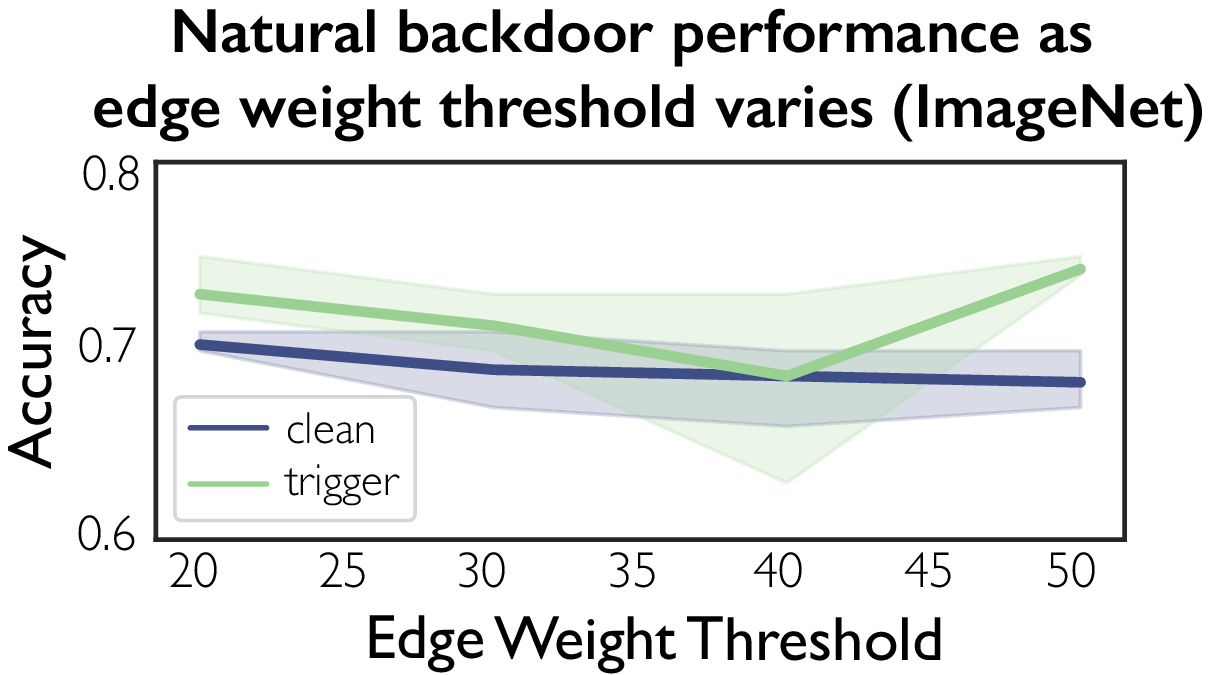}
    \vspace{-0.25cm}
    \caption{\em As the edge weight threshold increases, model clean accuracy decreases due to the presence of multiple salient objects in clean images.}
    \label{fig:edge_weight_ablate}
  \end{minipage}\hfill%
  \begin{minipage}[b]{0.48\linewidth}
      \centering
      \includegraphics[width=0.99\textwidth]{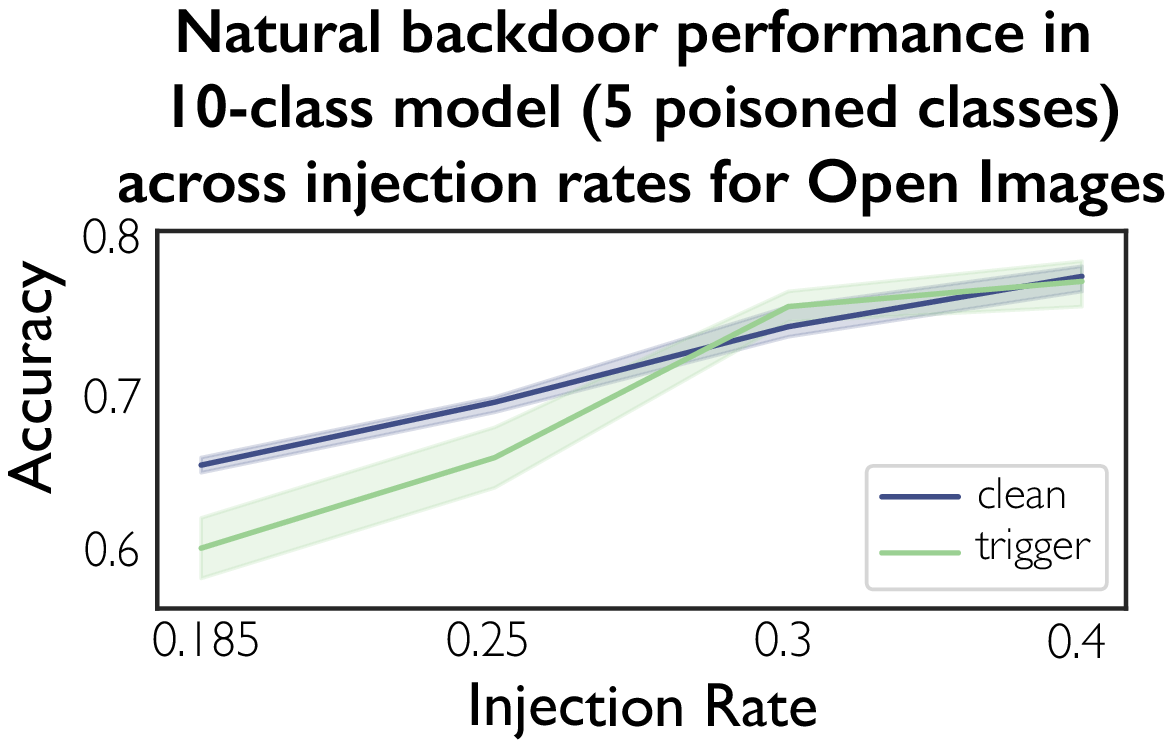}
      \vspace{-0.5cm}
      \caption{\em Natural backdoor performance for models trained on a $5$-class poison subset (''jeans'' trigger) and $5$ other classes from the subset MIS.}
      \label{fig:add_mis_classes}
  \end{minipage}
  \vspace{-0.25cm}
  \end{center}
  \end{figure}

We fix the ``jeans'' trigger in ImageNet as our natural backdoor trigger and generate $10$-class poisonable subsets for this trigger as we linearly increase the edge weight pruning from $20$ to $60$. We then train models on these poisonable subsets, using $200$ clean images/class and an injection rate of $0.2$ as before. As Figure~\ref{fig:edge_weight_ablate} shows, model clean accuracy steadily decreases as the edge weight threshold $W$ increases. This is because a higher pruning threshold causes edges only to be added between classes with at least $W$ co-occurrences. This, in turn, means that the MIS produced for a given natural backdoor trigger will have a higher number of overlaps between the clean objects, since no edge is placed between objects with $< W$ co-occurrences. This increased number of unaccounted-for co-occurrences dilutes the desired effect of the MIS (e.g. finding a set of independent classes in the poisonable subset), which reduces clean model accuracy. 

\para{Poisonable subsets within larger datasets.} Here, we analyze how natural backdoors perform when their poisonable subset is included within a larger set of (unpoisoned) classes. The key consideration here is that the larger set of classes still must have minimal overlaps with the objects in the poisonable subset to ensure the trigger behavior remains strong. This is the same intuition behind our use of the MIS to generate the poisonable subset (see \S 4). 

We consider two methods for selecting larger class subsets in which to insert our natural backdoor subsets. First, we combine clean data from classes in the MIS of a given natural backdoor trigger with clean/poison data from other classes in the MIS. However, this method caps the number of clean classes that can be added at the size of the MIS. Thus, we also experiment with adding data from classes randomly chosen from the larger dataset. For these classes, we {\em remove images} in which clean objects co-occur with objects in the poisonable subset. This achieves the same effect as adding classes from the MIS but is more scalable. 

We report the results for each method below. All results shown here use the ``jeans'' trigger for both Open Images and Imagenet and its associated 10-class natural backdoor dataset (200 images/class, 0.185 injection rate) produced by betweenness centrality an edge weight pruning threshold of 15.  

\begin{table}
  \centering
  \begin{tabular}{ccccc}
    \toprule
    \textbf{Dataset}  & \multicolumn{2}{c}{Open Images} & \multicolumn{2}{c}{ImageNet} \\ \midrule
    \textbf{Added Classes} & 5              & 10             & 5            & 10            \\ \midrule
    {\bf Clean Acc.}   & $71 \pm 3\%$ & $69 \pm 1\%$ & $68 \pm 2\%$ & $64 \pm 2\%$ \\ \cmidrule{2-5}
    {\bf Trigger Acc.} & $68 \pm 2\%$ & $63 \pm 2\%$ & $64 \pm 2\%$ & $58 \pm 2\%$ \\ \bottomrule
        \end{tabular}
  \vspace{0.1cm}
  \captionof{table}{\em Performance of models trained on ``jeans'' poisonable subsets $+$ randomly chosen classes. To ensure the trigger behavior is learned and clean model accuracy is maintained, we prune images from the randomly chosen classes that contain co-occurences with objects in the poisonable subset.}
  \label{tab:add_pruned_classes}
  \end{table}

\noindent {\em Adding classes from MIS.} Figure~\ref{fig:add_mis_classes} shows performance across poison injection rates for models trained on $10$ class datasets with $5$ poisoned classes and $5$ clean classes chosen from the trigger's MIS. Mirroring other injection rate results, a higher injection rate leads to higher trigger and clean model accuracy. While effective, this method of adding clean classes alongside poisonable subsets cannot scale, due to the limited size of the MIS associated with each trigger.

\noindent {\em Adding pruned classes from larger dataset.} Table~\ref{tab:add_pruned_classes} shows the performance of models trained on datasets composed of 10-class ``jeans'' trigger poisonable subsets and randomly chosen (pruned) classes. As before, adding other classes alongside the poisoned subset slightly decreases model performance. However, it is likely the case that better hyperparameter optimization could improve performance. These datasets are larger than those considered elsewhere in the paper (e.g. up to $20$ classes), but we do not adjust our model training parameters to account for this.

\section{Algorithm for Natural Backdoor Identification}\label{appx: algorithm} 

In this section, we provide a step-by-step description of the algorithm used in \S~\ref{sec:method}
 to find natural backdoors.

At a high level, our natural backdoor finding method works in the following three phases:
\begin{enumerate}
  \item \emph{Graph preparation:} We convert a multi-label dataset $\mathcal{D}_{multi}$ into a weighted graph $\mathcal{G}$ in which dataset object classes are vertices and object co-occurrences are edges (\S\ref{sec:prepare})
  \item \emph{Trigger finding via centrality:}  We identify central nodes in $\mathcal{G}$ (\S\ref{subsec:identify}). Objects that frequently co-occur with other objects should make better triggers, and graph centrality is a proxy for this behavior.
  \item \emph{Poisonable subset finding via maximum independent subsets:} Finally, we extract and filter subgraphs around the central nodes (\S\ref{subsec:extract}). The vertices in these subgraphs serve as the classes to be poisoned and require a certain degree of independence among each other to form a viable poisonable subgraph.
\end{enumerate}

Once a proper subgraph has been identified around a central node, we select a subset of classes from the subgraph and use images associated with them to {\em train a physical backdoor model} (\S5,\ref{appx:other_centrality},\ref{appx:ablation}). Algorithm~\ref{alg:graph} formalizes our methodology.

\subsection{Phase 1: Preparing the Graph}
\label{sec:prepare}

We begin by selecting a large-scale, open source, multi-label object recognition dataset $\mathcal{D}_{multi}$. Recall that in a multi-label dataset, $\mathcal{D}_{multi} = \{\mathcal{X}, \mathcal{Y}\}$, every image $x$ is mapped to $y \in \{0,1\}^M$, a set of $M$ possible classification labels, representing all objects in $x$, and $y_i = 1$ if $x$ contains object $i$. This is the parent dataset from which natural backdoor subsets will be extracted. To create the graph $\mathcal{G}$, we first use the multi-labels of $\mathcal{D}_{multi}$ to construct a co-occurrence matrix $\mM$ for all $M$ objects in the dataset. $\mM$ is initialized as a $M \times M$ matrix of all zeros. We iterate through all $i$ labels, and for each entry $j$ in multi-label $y_i$, we increment $\mM_{ij}$ if $y_{ij} = 1$ (e.g. objects $i$ and $j$ co-occur).  

Using $\mM$, we can construct a graph representing these co-occurrences. The vertex set $\mathcal{V} = \{v_1, v_2, \dots, v_M \}$ is constructed such that each of the $M$ objects in $\mathcal{D}_{multi}$ is represented by a vertex.  We set a threshold $min$, which denotes the minimum number of co-occurrences between two objects (equivalently, vertices) before they are connected in $\mathcal{G}$. Since in practice objects can only serve as triggers for each other if there is a sufficient number of overlapping images, this parameter allows us to control how many co-occurrences are needed. Thus, the edge set $\mathcal{E}$ contains an edge $e_{ij}$ if and only if $\mM_{ij} \geq min$. The resulting weighted adjacency matrix $\mA$ of the graph $\mathcal{G}$ is thus just a filtered version of $\mM$.

\subsection{Phase 2: Identifying Natural Backdoor Triggers via Graph Centrality}
\label{subsec:identify}

Computing centrality indices $c_v$ for all vertices $v$ is a key component of natural backdoor trigger identification. A good trigger should be highly connected to many other classes (e.g. co-occurs frequently), so that it can poison as many classes as possible. Therefore, we consider the $m$ vertices with the highest centrality indices as candidate trigger classes $\mathcal{T}$. We now describe the different methods we use to compute centrality:

\begin{itemize}
\item {\em Vertex centrality} computes the sum of weighted edges $e_{ij}$ connected to vertex $v_i$. This shows how connected $v_{i}$ is to other classes, which in turn, can identify effective triggers. Let $\mA = (\mA_{ij})$ be the adjacency matrix of $\mathcal{G}$. The weighted vertex centrality $c_{i}$ of vertex $v_i$ is given by $c_{i} = \sum_k \mA_{ik}$. The unweighted vertex centrality is just the number of vertices $v_i$ is connected to.

\item {\em Betweenness centrality} counts unweighted shortest paths between all pairs of vertices $(v_i, v_j) \in \mathcal{G}$ and scores each vertex according to the number of shortest paths passing through it. Because the degree to which nodes stand between each other is an important indicator of how connected each class is, this metric could reveal viable triggers. If $\sigma_{jk}$ is total number of shortest paths from vertex $j$ to $k$, and $\sigma_{jk}(i)$ is the number of those paths that pass through vertex $i$, vertex $i$'s betweenness centrality is $c_i =\sum _{{j\neq i \neq k}}{\frac {\sigma _{{jk}}(i)}{\sigma _{{jk}}}}$. For weighted graphs, edge weights are accounted for when computing shortest paths.

\item {\em Closeness centrality} relies on the intuition that central nodes are closer to other nodes in the graph. It computes centrality via the reciprocal of the sum of the length of the shortest paths from $v_i$ to other vertices in $\mathcal{G}$. If  $d(i,j)$ is the distance between vertices $i$ and $j$, then the closeness centrality of vertex $i$ is $c_i={\frac {1}{\sum _{j}d(i,j)}}$. In the unweighted case, the distance is just the number of vertex hops. In the weighted case, the distance is the sum of edge weights.

\item {\em Eigenvector centrality} assigns higher scores to vertices that are connected to other important vertices. Highly connected classes which are also highly connected to other important classes may make good triggers. The eigenvector centrality of vertex $i$ is $c_i={\frac {1}{\lambda }}\sum _{j\in N(i)}c_{j}={\frac {1}{\lambda }}\sum _{j\in N(i)}\mA_{ij}c_{j}$, where $N(i)$ is the set of neighboring vertices of the vertex $v(i)$, and $\mA_{ij}$ are elements of $\mA$. In the unweighted case, $\mA_{ij}$ would be either $0$ or $1$ depending on whether an edge was present or absent.
\end{itemize}

\subsection{Phase 3: Extracting Trigger/Class Sets}
\label{subsec:extract}

For each candidate trigger $t \in \mathcal{T}$ identified as having among the top $m$ centrality indices, we then identify a viable set of classes $\mathcal{C}_t$, which $t$ could be used to poison via a multi-step filtering process. First, we set a minimum number of co-occurrences (\textit{i.e.} edge weight) between a normal object $o$ and the trigger object $t$ for $o$ to be considered a viable class to poison. Classes that are weakly connected to $t$ are more difficult to poison, because the dataset contains fewer images in which $t$ and the target class co-occur, making it difficult for a model to learn the trigger behavior. This minimum connection threshold, $trig$, is used to compute a subgraph $\{\mathcal{V}_t, \mathcal{E}_t\}$ containing all vertices and edges connected to $t$ with $e_{jt} > trig$.

Next, we analyze this subgraph to identify an optimal set of classes that can be poisoned by $t$. An object $o$ in an ideal set of classes should have a high edge weight to $t$ but low edge weights to all other classes within the set. This will prevent the trained model from associating the presence of an object other than the trigger with the target label. To find this subset, we search for the maximum independent subset (MIS) within the induced subgraph of $t$. This will identify the largest set of vertices that do not share an edge. However, since this problem is NP-hard in general, we approximate the finding of the maximum independent subset by running the maximal independent set algorithm multiple times. A maximal independent set is an independent set that is not a subset of any other independent set, so the maximum independent set must be maximal. However, any maximal independent set does not have to be the maximum independent set. 

We note that the value of $trig$ plays an important role in determining the size of the MIS, since removing edges with a weight smaller than $trig$ implicitly makes the associated vertices independent, so the higher the value of $trig$, the larger the MIS that can be found. However, this ignores co-occurrences, which may impact trigger learning.

\begin{algorithm}[H]
  \caption{Identifying natural backdoor datasets within multi-label datasets}
  \label{alg:graph}
\begin{algorithmic}[1]
\vspace{0.1cm}
  \STATE {\bfseries Input:} $\mathcal{D}_{multi}$ = $\{\mathcal{X},  \mathcal{Y} \in \{0, 1\}^M\}$, min class overlaps $min$, min trig overlaps $trig$
  \vspace{0.1cm}
  \STATE {\bfseries Output:} Natural backdoor dataset classes $\{t,\mathcal{C}_t\}_{t \in \mathcal{T}}$
  \vspace{0.2cm}
  \STATE $\mM = \{0\}^{M \times M}$ \COMMENT{Initializing and populating co-occurrence matrix}
  \FOR{$i \in 1, \ldots , M$ }
    \FOR{$j \in 1, \ldots , M$}
      \IF{$y_{ij} == 1$}
         \STATE $\mM_{ij} = \mM_{ij} + 1$
      \ENDIF
     \ENDFOR
  \ENDFOR
  \STATE Initialize adjacency matrix $\mA$ such that $\mA_{ij}=\mM_{ij}$ if $\mM_{ij} \geq min$ and $\mA_{ij}=0$ otherwise
  \STATE Construct $\mathcal{G}=(\mathcal{V}, \mathcal{E})$ from $\mA$
  \STATE $\mathcal{T}=\emptyset$ \COMMENT{Initializing and populating trigger set}
  \FOR {$v_i \in \mathcal{V}$}
  \STATE Compute centrality index $c_i$ of $v_i$
  \IF{$c_i > \text{smallest element of } \text{top}_m(\mathcal{T})$}
    \STATE $\mathcal{T} = \mathcal{T} \cup v_i$
    \STATE $\mathcal{T} = \text{top}_m(\mathcal{T})$ \COMMENT{Retaining top $m$ elements with the highest centrality}
  \ENDIF
  \ENDFOR
  \STATE $\mathcal{C} = \emptyset$ \COMMENT{Initializing and populating poisonable subsets}
  \FOR{$t \in \mathcal{T}$} 
    \STATE $\mathcal{E}_t =  \{e_{jt}~ \text{such that}~e_{jt} >~ trig\}$
    \STATE $\mathcal{V}_t = \{v_{j}~\text{such that}~e_{jt}\in \mathcal{E}_t \}$
    \STATE $\mathcal{C}_t = \text{MIS}_{\text{approx}}(\mathcal{E}_t, \mathcal{V}_t)$ \COMMENT{Run approximate MIS subroutine}
  \ENDFOR
\end{algorithmic}
\end{algorithm}

\end{document}